\pdfoutput=1

\documentclass[11pt,table,xcdraw]{article}

\usepackage{acl}
\usepackage{kotex}
\usepackage{tabularx}
\usepackage{booktabs}
\usepackage{boldline}
\usepackage{subcaption}
\usepackage{multirow}
\usepackage{float}
\usepackage{xcolor}

\usepackage{times}
\usepackage{latexsym}

\usepackage{graphicx}

\usepackage[T1]{fontenc}

\usepackage[utf8]{inputenc}

\usepackage{microtype}

\title{Mismatch between Multi-turn Dialogue \\
and its Evaluation Metric in Dialogue State Tracking}

\author{Takyoung Kim$^1$, Hoonsang Yoon$^1$, Yukyung Lee$^1$, Pilsung Kang$^1$, Misuk Kim$^2$ \\
        $^1$Korea University, Seoul, Republic of Korea \\ 
        $^2$Sejong University, Seoul, Republic of Korea \\
        \texttt{$^1$\{takyoung\_kim, hoonsang\_yoon, yukyung\_lee, pilsung\_kang\}@korea.ac.kr}\\ 
        \texttt{$^2$misuk.kim@sejong.ac.kr}}

\begin{document}
\maketitle
\begin{abstract}
Dialogue state tracking (DST) aims to extract essential information from multi-turn dialogue situations and take appropriate actions. A belief state, one of the core pieces of information, refers to the subject and its specific content, and appears in the form of \texttt{domain-slot-value}. The trained model predicts ``accumulated'' belief states in every turn, and joint goal accuracy and slot accuracy are mainly used to evaluate the prediction; however, we specify that the current evaluation metrics have a critical limitation when evaluating belief states accumulated as the dialogue proceeds, especially in the most used MultiWOZ dataset. Additionally, we propose \textbf{relative slot accuracy} to complement existing metrics. Relative slot accuracy does not depend on the number of predefined slots, and allows intuitive evaluation by assigning relative scores according to the turn of each dialogue. This study  also encourages not solely  the reporting of joint goal accuracy, but also various complementary metrics in DST tasks for the sake of a realistic evaluation.
\end{abstract}


\section{Introduction}
The dialogue state tracking (DST) module structures the belief state that appears during the conversation in the form of \texttt{domain-slot-value}, to provide an appropriate response to the user. Recently, multi-turn DST datasets have been constructed using the Wizard-of-Oz method to reflect more realistic dialogue situations \citep{wen-etal-2017-network, mrksic-etal-2017-neural, budzianowski-etal-2018-multiwoz}. The characteristic of these datasets is that belief states are ``accumulated'' and recorded every turn. That is, the belief states of the previous turns are included in the current turn. It confirms whether the DST model tracks essential information that has appeared up to the present point.

Joint goal accuracy and slot accuracy are utilized  in most cases to evaluate the prediction of accumulated belief states. Joint goal accuracy strictly determines whether every predicted state is identical to the gold state, whereas slot accuracy measures the ratio of correct predictions. However, we determined  that these two metrics solely  focus on ``penalizing states that fail to predict,'' not considering ``reward for well-predicted states.'' Accordingly, as also pointed out in \citet{Rastogi2020SchemaGuidedDS}, joint goal accuracy underestimates the model prediction because of its error accumulation attribute, while slot accuracy overestimates it because of its dependency on predefined slots.

However, there is a lack of discussion on the metric for evaluating the most used MultiWOZ dataset, despite a recently published dataset \citep{rastogi2020towards} proposing some metrics. To address the above challenge, we propose reporting the \textbf{relative slot accuracy} along with the existing metrics in MultiWOZ dataset. While slot accuracy has the challenge of overestimation by always considering all predefined slots in every turn, relative slot accuracy does not depend on predefined slots, and calculates a score that is affected solely  by slots that appear in the current dialogue. Therefore, relative slot accuracy enables a realistic evaluation by rewarding the model's correct predictions, a complementary approach that joint goal  and slot accuracies  cannot fully cover. It is expected that the proposed metric can be adopted  to evaluate model performance more intuitively.


\section{Current Evaluation Metrics}
\label{sec:current}

\subsection{Joint Goal Accuracy}
\label{sec:jga}
Joint goal accuracy, developed from \citet{henderson-etal-2014-word} and \citet{zhong-etal-2018-global}, can be said to be an ideal metric, in that it verifies that the predicted belief states perfectly match the gold label. Equation \ref{eq:jga} expresses  how to calculate the joint goal accuracy, depending on whether the slot values match each turn.

\begin{equation}
\label{eq:jga}
JGA = \left\{ 
  \begin{array}{ c l }
    1 & \, \textrm{if predicted state}=\textrm{gold state}  \\
    0 & \, \textrm{otherwise}
  \end{array}
\right.
\end{equation}

\begin{figure}[t!]
    \centering
    \includegraphics[width=\columnwidth]{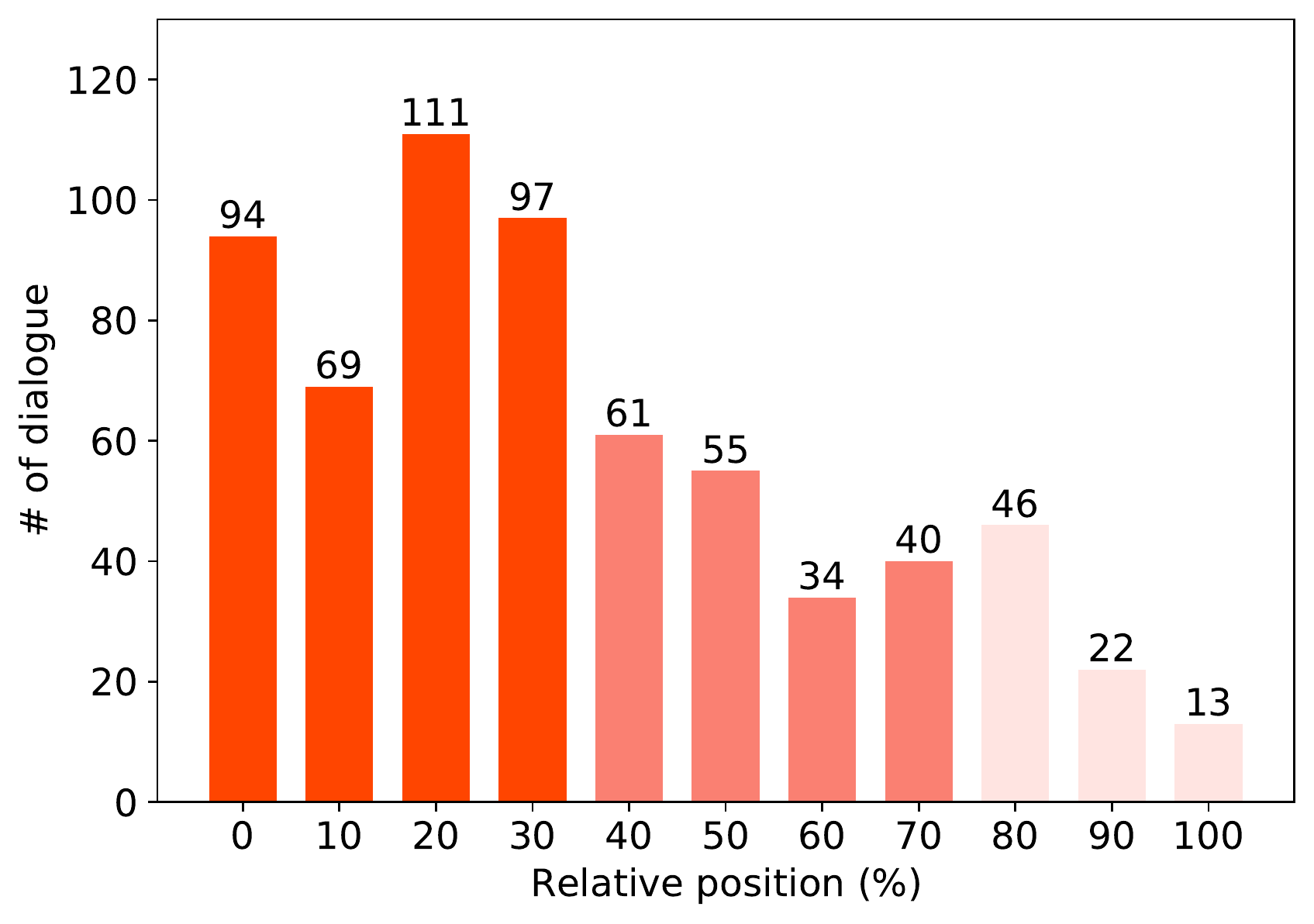}
    \caption
        {
        The relative position where joint goal accuracy of the turn is measured to be zero for the first time among the dialogues where joint goal accuracy of the last turn is zero. (642 of 999 MultiWOZ 2.1 test set with SOM-DST).
        }
    \label{fig:jga}
\end{figure}

However, the joint goal accuracy underestimates the accumulated states because it scores the performances of later turn to zero if the model mispredicts even once in a particular turn, regardless of the model prediction quality at later turns. As illustrated  in Figure \ref{fig:jga}, we measured the relative position of the turn causing this phenomenon for the dialogue. We used MultiWOZ 2.1 \citep{DBLP:journals/corr/abs-1907-01669}, and analyzed 642 samples from a total of 999 test sets in which the joint goal accuracy of the last turn is zero. The DST model selected for primary verification is the SOM-DST \citep{kim-etal-2020-efficient}, which is one of the latest DST models. Accordingly, the relative position where joint goal accuracy first became zero was mainly at the beginning of the dialogue\footnote{59  samples of the 642 samples have a joint goal accuracy of 1 in the middle, owing  to a coincidental situation or differences in the analysis of annotation. Table \ref{table:dialog_1} and Table \ref{table:dialog_2} show the dialogue situation in detail, and Table \ref{table:dialog_state_1} and Table \ref{table:dialog_state_2} show the belief states accordingly. Refer to Appendix \ref{sec:appendix_a}.}. This means that the joint goal accuracy after the beginning of the dialogue is unconditionally measured as zero because of the initial misprediction, although the model may correctly predict new belief states at later turns. Failure to measure the performance of the latter part means that it cannot consider various dialogue situations provided  in the dataset, which is a critical issue in building a realistic DST model.

\begin{figure}[t!]
    \centering
    \includegraphics[width=\columnwidth]{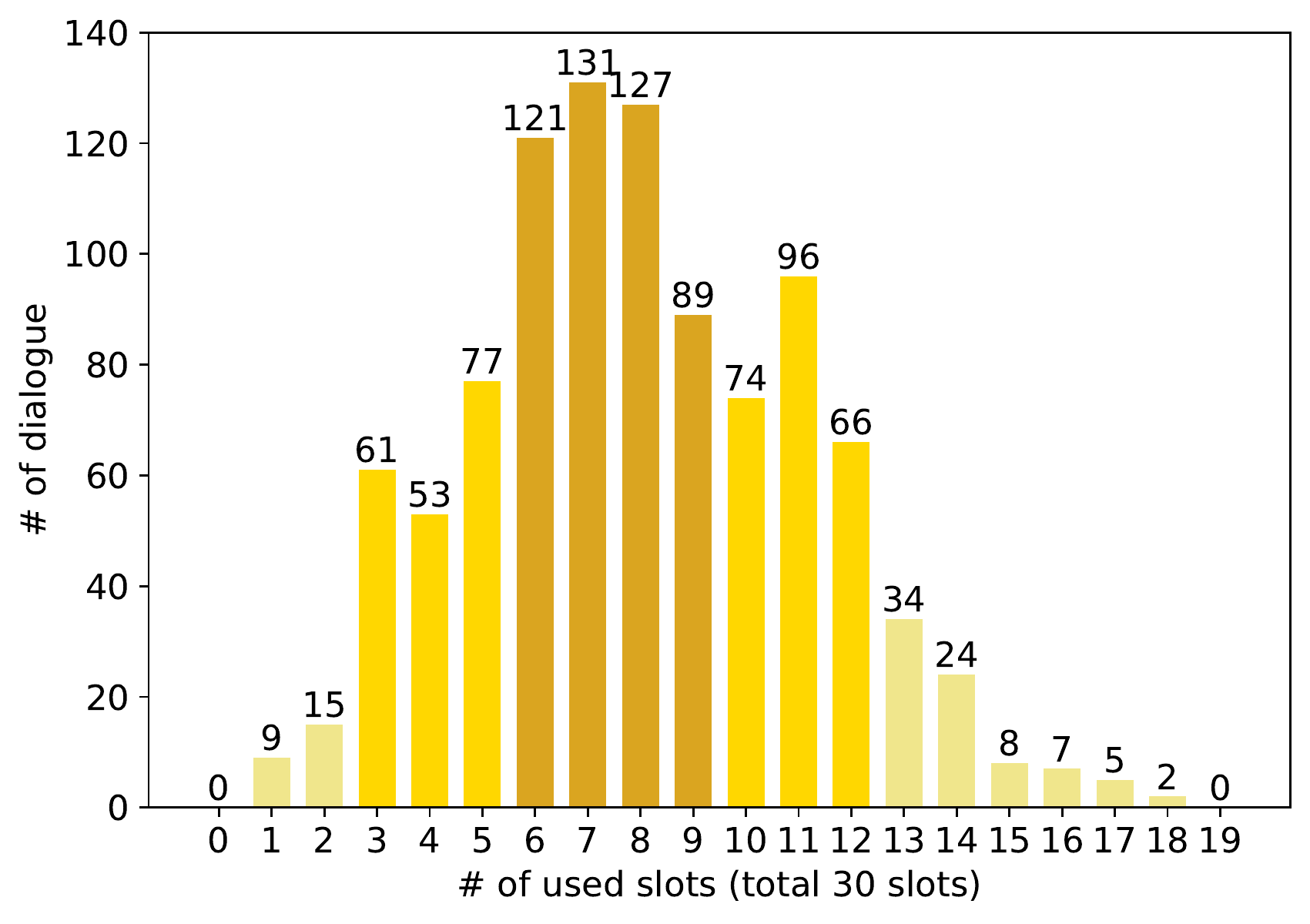}
    \caption
        {
        The number of predefined gold slots used in each dialogue (999 MultiWOZ 2.1 test set).
        }
    \label{fig:sa}
\end{figure}

\begin{table*}[t!]
\centering 
\renewcommand{\arraystretch}{1.0}
\setlength{\tabcolsep}{12pt}
\begin{tabular}{llcccc}
\toprule
\multirow{2}{*}{\textbf{Type}} & \multirow{2}{*}{\textbf{Model}} &  \textbf{Joint}   &  \textbf{Slot} &  \textbf{F1} & \textbf{Relative}  \\ 
& & \textbf{Goal Acc.} & \textbf{Acc.} & \textbf{Score} & \textbf{Slot Acc.} \\
\midrule
 \multirow{7}{*}{\shortstack[l]{Open\\vocabulary}}  &  Transformer-DST \citeyearpar{zeng2021jointly} &   0.5446 &   0.9748  &  0.9229 & 0.8759 \\
 &  TripPy \citeyearpar{heck-etal-2020-trippy} &   \textbf{0.6131} &   0.9707  &   0.8573 &   0.8432 \\
 &  SOM-DST \citeyearpar{kim-etal-2020-efficient} &   0.5242 &   0.9735  &   0.9179 &   0.8695 \\
 &  Simple-TOD \citeyearpar{NEURIPS2020_e9462095} &   0.5605 &   \textbf{0.9761} &   \textbf{0.9276} &   \textbf{0.8797}  \\
 &  SAVN \citeyearpar{wang-etal-2020-slot} &   0.5357 &   0.9749  &   0.9246 &   0.8769 \\
 &  TRADE \citeyearpar{wu-etal-2019-transferable} &   0.4939 &   0.9700  &   0.9033 &   0.8520 \\ 
 &  COMER \citeyearpar{ren-etal-2019-scalable} &   0.4879 &   0.9652 &   0.8800 &   0.8250 \\ \hline 
 \multirow{3}{*}{\shortstack[l]{Ontology\\based}} &  DST-STAR \citeyearpar{10.1145/3442381.3449939} &   0.5483 &   0.9754 &   0.9253 &   0.8780  \\
 &  L4P4K2-DSGraph \citeyearpar{lin2021knowledgeaware} &   0.5178 &   0.9690 &   0.9189  &   0.8570\\
 &  SUMBT \citeyearpar{lee-etal-2019-sumbt} &   0.4699 &   0.9666 &   0.8934 &   0.8380 \\
\bottomrule
\end{tabular}
\caption{\label{table:result}
Model performance of MultiWOZ 2.1 with various evaluation metrics. All reported performances are our re-implementation.
}
\end{table*}

\subsection{Slot Accuracy}
\label{sec:sa}

Slot accuracy can compensate for situations where joint goal accuracy does not fully evaluate the dialogue situation. Equation \ref{eq:sa} expresses how to calculate the slot accuracy. $T$ indicates the total number of predefined slots for all the domains. $M$ denotes the number of missed slots that the model does not accurately predict among the slots included in the gold state, and $W$ denotes the number of wrongly predicted  slots among the slots that do not exist in the gold state.

\begin{equation}
\label{eq:sa}
    SA = {T-M-W \over T}
\end{equation}

Figure \ref{fig:sa} illustrates the total number of annotated slots in MultiWOZ 2.1 to figure out the limitation of slot accuracy. Each value of $x$-axis in Figure \ref{fig:sa} indicates the “maximum” number of slots that appear  in a single dialogue, and we confirmed that approximately 85\% of the test set utilized  solely  less than 12 of the 30 predefined slots in the experiment. Because the number of belief states appearing in the early and middle turns of the dialogue are smaller, and even fewer states make false predictions, calculating slot accuracy using Equation \ref{eq:sa} reduces the influence of $M$ and $W$, and the final score is dominated by the total slot number $T$. Accordingly, several  previous studies still report the model performance using solely  joint goal accuracy because slot accuracy excessively depends on the number of predefined slots, making the performance deviation among models trivial (refer to  Table \ref{table:survey}).

Furthermore, according to Table \ref{table:extreme}, we determined  that slot accuracy tends to be too high. The slot accuracies  of turns 0 and 1 show approximately 96\% accuracy, despite the model not correctly predicting states at all. It becomes difficult to compare various models in detail, if each model shows a high performance, even though nothing is adequately predicted. In addition, as the turn progresses, there are no rewards for a  situation in which the model tracks the belief state without any challenges. The case correctly predicting two out of three in turn 4, and the case correctly predicting three out of four in turn 5 exhibit the same slot accuracy. Therefore, the slot accuracy measured according to Equation \ref{eq:sa} differs from our intuition.

\subsection{Other Metric}
\label{sec:aga}
Recently, \citet{rastogi2020towards} proposed a metric called average goal accuracy. The main difference between the average goal accuracy and the proposed relative slot accuracy is that the average goal accuracy only considers the slots with non-empty values in the gold states of each turn, whereas the proposed relative slot accuracy considers those in both gold and predicted states. Since average goal accuracy ignores the predicted states, it cannot properly distinguish a better model from a worse model in some specific situations. We will discuss it in more detail in Section \ref{sec:discussion}.


\section{Relative Slot Accuracy}

As can be observed  in Equation \ref{eq:sa}, slot accuracy has the characteristic that the larger the number of predefined slots ($T$), the smaller the deviation between the prediction results. The deviation among DST models will be even more minor when constructing datasets with various dialogue situations, because the number of predefined slots will continually increase. It is not  presumed  to be an appropriate metric in terms of scalability.

Therefore, we propose relative slot accuracy, that is not affected by predefined slots, and is evaluated  with adequate rewards and penalties that fit human intuition in every turn. Equation \ref{eq:rsa} expresses  how to calculate the relative slot accuracy, and $T^\ast$ denotes the number of unique slots appearing in the predicted  and gold states in a particular turn.

\begin{equation}
\label{eq:rsa}
    RSA = {T^\ast - M -W \over T^\ast} \mbox{, where $0$ if $T^\ast=0$}
\end{equation}

Relative slot accuracy rewards well-predicted belief states by measuring the scores in accumulating turns. Further discussions on the relative score will be discussed in Section \ref{sec:discussion}.


\section{Experiments}

We measured MultiWOZ 2.1, an improved version of MultiWOZ 2.0 \citep{budzianowski-etal-2018-multiwoz}, which has been adopted in several  studies, according to Table \ref{table:survey}. Five domains (i.e., \textit{hotel, train, restaurant, attraction,} and \textit{taxi}) are adopted  in the experiment, following \citet{wu-etal-2019-transferable}, and there are a total of 30 \texttt{domain-slot} pairs. We selected the DST models in Table \ref{table:survey} that perform the MultiWOZ experiment with the original authors' reproducible code\footnote{Implementation codes for Simple-TOD and TripPy are from https://github.com/salesforce/coco-dst.}. Additionally, we reported the F1 score, which can be calculated using the current predicted  and  gold states.

\begin{figure}[ht]
    \centering
    \includegraphics[width=0.9\columnwidth]{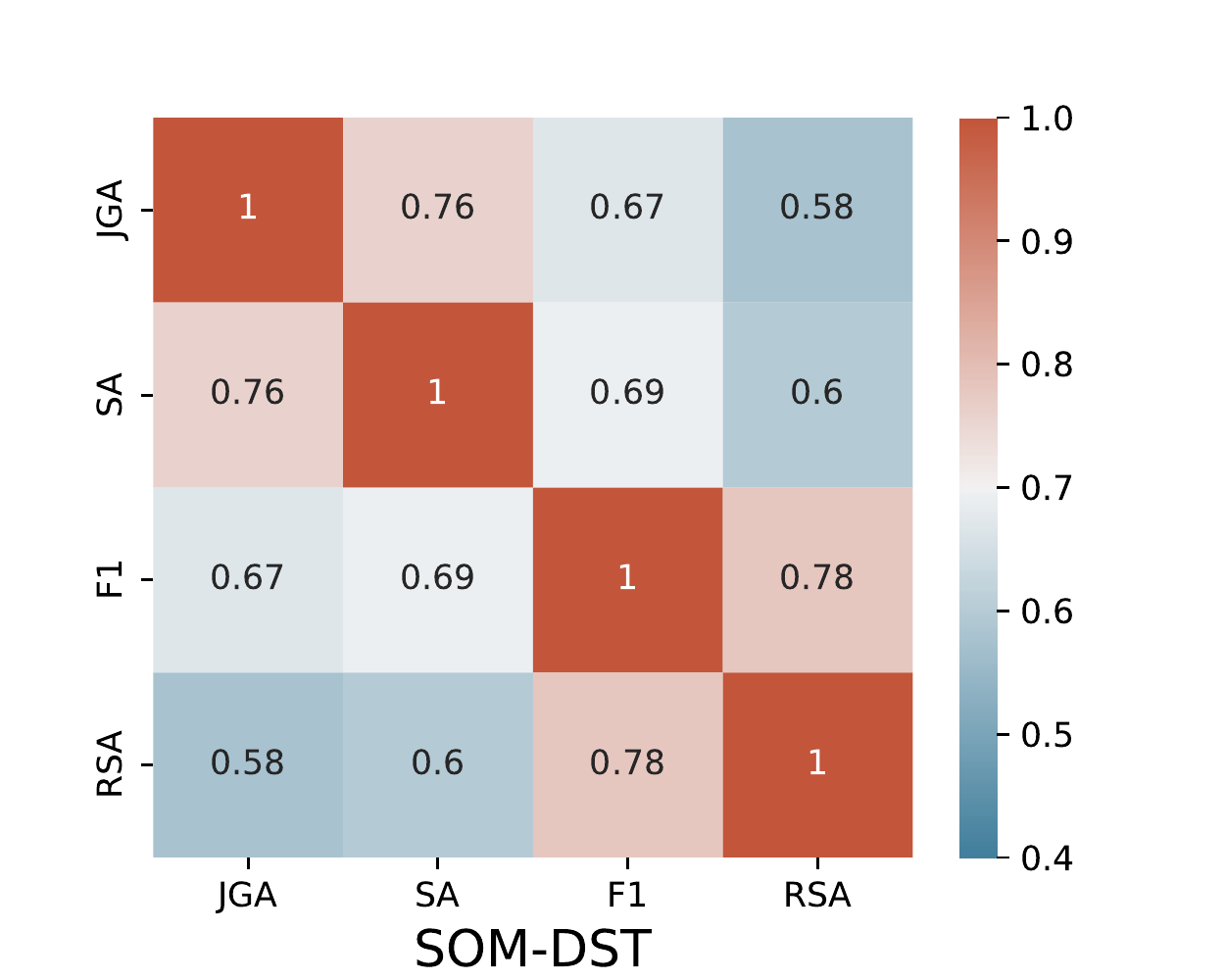}
    \caption
        {
        Correlation matrix of evaluation performance of total 7,368 turns in 999 MultiWOZ 2.1 test set using SOM-DST. Results for other models are included in Figure \ref{fig:corrs}.}
    \label{fig:corr}
\end{figure}

\subsection{Results and Discussion}
\label{sec:discussion}

Table \ref{table:result} presents  the overall results. Regarding  slot accuracy, the difference between the largest and smallest values is solely  1.09\%. It can be one of the reasons that several researchers do not report it. Meanwhile, relative slot accuracy can explicitly highlight the deviation among models by showing a 5.47\% difference between the largest and smallest values. Furthermore, the correlation with joint goal accuracy, a mainly adopted metric, and relative slot accuracy with respect to each turn is lower than the correlation with joint goal accuracy and slot accuracy, as illustrated in Figure \ref{fig:corr}. Specifically, it can be compared with a different perspective when using the proposed reward-considering evaluation metric.

\paragraph{Domain-specific Evaluation} 

We reported the joint goal, slot, and relative slot accuracies  per domain utilizing the SOM-DST model in Table \ref{tab:perdomain}. Relative slot accuracy derives a specific score in the turn configuration and prediction ratio of each domain by excluding slots that do not appear in the conversation. For example, the \textit{taxi} domain shows a low score, meaning that it has relatively several cases of incorrect predictions, compared to the number of times  slots belonging to the \textit{taxi} domain appear. Because slot accuracy cannot distinguish the above trend, the score of the \textit{hotel} domain is lower than that of the \textit{taxi} domain. In summary, relative slot accuracy enables relative comparison according to the distribution of the domain in a dialogue.

\begin{table}[t]
\centering 
\renewcommand{\arraystretch}{1.0}
\setlength{\tabcolsep}{7pt}
\begin{tabular*}{\columnwidth}{ccccc} 
\hlineB{3}
\multirow{2}{*}{ \textbf{Domain}}   &  \textbf{Joint} &  \textbf{Slot} &  \textbf{Relative} \\ 
&  \textbf{Goal Acc.} &  \textbf{Acc.} &  \textbf{Slot Acc.} \\ \hlineB{1}
   hotel &   0.4923 &   0.9731 &    0.8493 \\
   train &   0.7162 &   0.9874&    0.9176 \\
   restaurant &   0.6589 &    0.9858 &   0.8977 \\
   attraction &   0.6811 &   0.9878 &   0.8421 \\
   taxi &   0.5701 &   0.9798 &   0.7828 \\
\hlineB{3}
\end{tabular*}
\caption
{
Per-domain performance of SOM-DST prediction.
}
\label{tab:perdomain}
\end{table}

\begin{figure}[t]
    \centering
    \includegraphics[width=0.8\columnwidth]{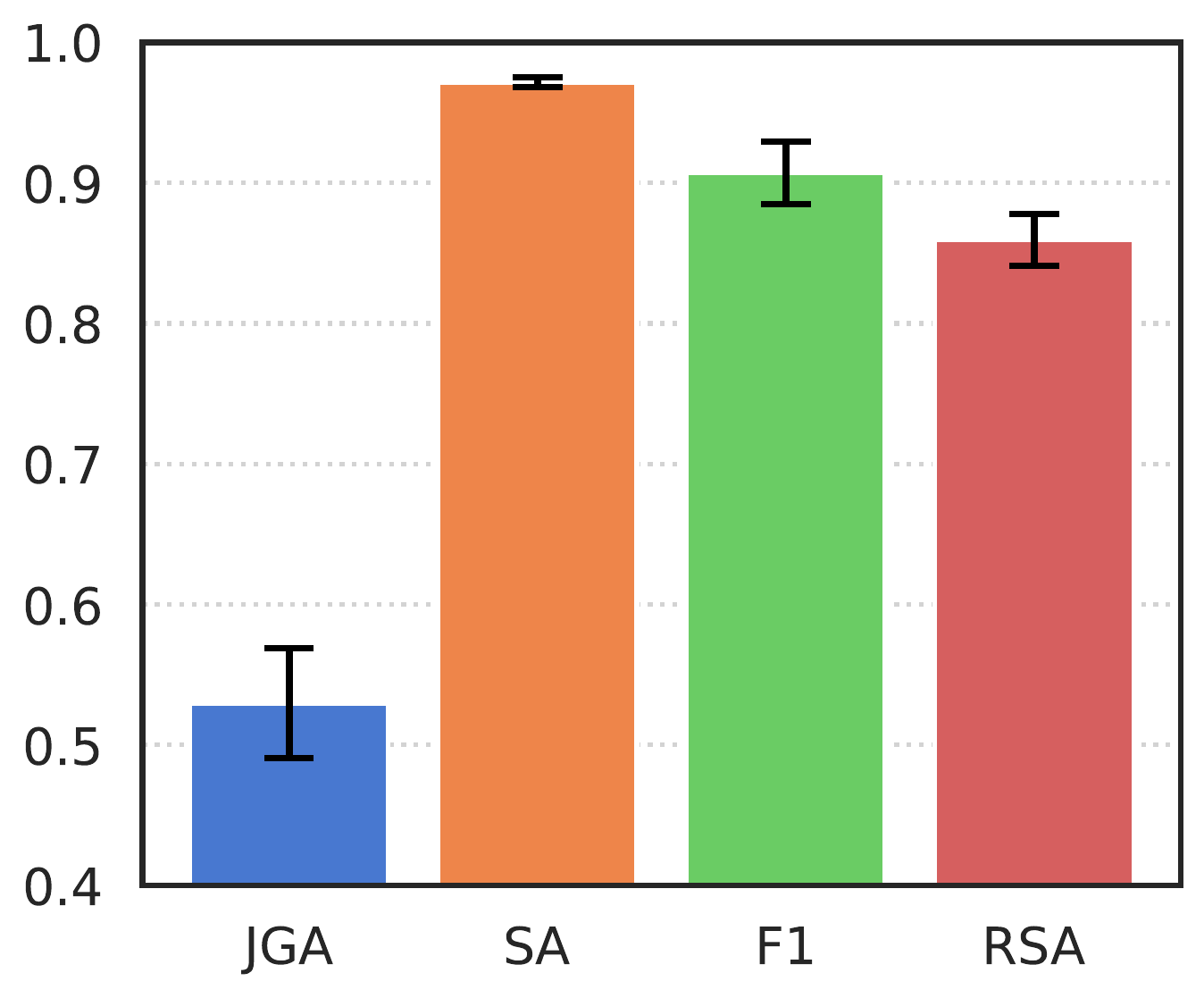}
    \caption
        {
        The mean and standard deviation of model performance reported in Table \ref{table:result}.
        }
    \label{fig:deviation}
\end{figure}

\begin{table*}[t!]
\centering
\renewcommand{\arraystretch}{0.8}
\setlength{\tabcolsep}{10pt}
\begin{tabular}{llccc}
\toprule
\multirow{2}{*}{\textbf{Type}} & \multirow{2}{*}{\textbf{Belief State}} &  \textbf{Joint}   &  \textbf{Average} &  \textbf{Relative} \\ 
& & \textbf{Goal Acc.} & \textbf{Goal Acc.} & \textbf{Slot Acc.} \\
\midrule
 \multirow{3}{*}{\shortstack[l]{Gold State}}  &  \textcolor{blue}{restaurant-area-centre} & \multirow{3}{*}{\shortstack[l]{-}}  & \multirow{3}{*}{\shortstack[l]{-}} & \multirow{3}{*}{\shortstack[l]{-}}  \\
 &  \textcolor{orange}{restaurant-food-indian} &   &   &  \\
 &  \textcolor{orange}{restaurant-people-2} &   &   &  \\
 \hline
 \multirow{3}{*}{\shortstack[l]{Prediction\\of Model A}}  &  \textcolor{blue}{restaurant-area-centre} & \multirow{3}{*}{\shortstack[l]{0}}  & \multirow{3}{*}{\shortstack[l]{0.3333}} & \multirow{3}{*}{\shortstack[l]{0.2500}}  \\
 &  \textcolor{orange}{restaurant-food-chinese} &   &   &  \\
 &  \textcolor{magenta}{attraction-area-centre} &   &   &  \\
 \hline
 \multirow{5}{*}{\shortstack[l]{Prediction \\ of Model B}}  &  \textcolor{blue}{restaurant-area-centre} & \multirow{5}{*}{\shortstack[l]{0}} & \multirow{5}{*}{\shortstack[l]{0.3333}} & \multirow{5}{*}{\shortstack[l]{0.1667}}  \\
 &  \textcolor{orange}{restaurant-food-chinese} &   &   &  \\
 &  \textcolor{magenta}{restaurant-name-nusha} &   &   &  \\
 &  \textcolor{magenta}{attraction-area-centre} &   &   &  \\
 &  \textcolor{magenta}{attraction-pricerange-cheap} &   &   &  \\
\bottomrule
\end{tabular}
\caption{\label{table:aga}
A situation that average goal accuracy cannot distinguish between two models. States with \textcolor{blue}{blue} denote correct prediction, and as defined in Section \ref{sec:sa}, states with \textcolor{orange}{orange} and \textcolor{magenta}{pink} denote respective $M$ and $W$.
}
\end{table*}

\paragraph{Dependency on Predefined Slots} 
As discussed in Section \ref{sec:sa}, slot accuracy requiring total predefined slots is not a scalable method for evaluating the current dialogue dataset that contains a few domains in each dialogue. For example, when evaluating a dialogue sample that solely deals with the \textit{restaurant} domain, even domains that never appear at all (i.e., \textit{hotel, train, attraction,} and \textit{taxi}) are involved in measuring performance, making deviations among different models trivial. However, relative slot accuracy can evaluate the model's predictive score without being affected by slots never seen in the current dialogue, which is a more realistic way, considering that each dialogue contains its own turn and slot composition. Figure \ref{fig:deviation} illustrates the mean and standard deviations of the model performance in Table \ref{table:result}. As can be observed from the results, the relative slot accuracy has a higher deviation than the slot accuracy, enabling a detailed comparison among the methodologies.

\paragraph{Reward on Relative Dialogue Turn}  
Relative slot accuracy is able to reward the model's correct prediction by measuring the accuracy on a relative basis for each turn. Table \ref{table:extreme} compares the slot and relative slot accuracies. The relative slot accuracy from turns 0 -- 3 is measured as 0 because it calculates the score based on the unique state of the current turn according to Equation \ref{eq:rsa}. In addition, regarding slot accuracy in turns 4, 5, and 6, there is no score improvement for the additional well-predicted state by the model, whereas the score increases when the newly added state is matched in the case of relative slot accuracy. Therefore, relative slot accuracy can provide an intuitive evaluation reflecting the current belief state recording method, in which the number of slots accumulates incrementally as the conversation progresses.

\paragraph{Comparison to Average Goal Accuracy}
Relative slot accuracy can compare DST model performances more properly than average goal accuracy, as mentioned in Section \ref{sec:aga}. Table \ref{table:aga} describes how these two metrics result in different values for the same model predictions. In this example, average goal accuracy cannot consider additional belief states incorrectly predicted by \texttt{Model B}, resulting in the same score between the two models. In contrast, relative slot accuracy can give a penalty proportional to the number of wrong predictions because it includes both gold and predicted states when calculating the score. Consequently, relative slot accuracy has a more elaborated discriminative power than the average goal accuracy.


\section{Conclusion}
This paper points out the challenge that the existing joint goal and slot accuracies cannot fully evaluate the accumulating belief state of each turn in the MultiWOZ dataset. Accordingly, the relative slot accuracy is proposed. This metric is not affected by unseen slots in the current dialogue situation, and compensates for the model's correct prediction. When the DST task is scaled up to deal with more diverse conversational situations, a realistic model evaluation will be possible using relative slot accuracy. Moreover, we suggest reporting various evaluation metrics to complement the limitations of each metric in future studies, not solely reporting the joint goal accuracy.

\section*{Acknowledgement}
This work was supported by Institute of Information \& communications Technology Planning \& Evaluation (IITP) grant funded by the Korea government (MSIT) (No. 2021-0-00034, Clustering technologies of fragmented data for time-based data analysis) and Ministry of Culture, Sports and Tourism and Korea Creative Content Agency (Project Number: R2020040126-0001)

\bibliography{anthology,custom}
\bibliographystyle{acl_natbib}

\appendix
\section{Complementary discussions of joint goal accuracy}
\label{sec:appendix_a}

Our findings show that if the model makes an incorrect prediction, the error accumulates until the end of the dialogue, and the joint goal accuracy remains at zero. In this section, we discuss a few cases of 59 dialogues that do not show the trend among 642 dialogues selected in Section \ref{sec:jga}; however, it is important to note that these few cases have negligible effect on the trend in Figure \ref{fig:jga}, solely changing the position where the joint goal accuracy first becomes zero.

We sampled dialogues of the MultiWOZ 2.1 test set in Table \ref{table:dialog_1} and Table \ref{table:dialog_2}, and marked values appearing in the dialogue in bold. Table \ref{table:dialog_state_1} and Table \ref{table:dialog_state_2} indicate the corresponding belief states of each dialogue. In the first dialogue presented in Table \ref{table:dialog_1}, the joint goal accuracy is measured as 1 at turn 2. In this case, the model incorrectly predicted the \texttt{restaurant-pricerange} slot at turns 0 and 1, and then the utterance about the slot appeared by chance. In a general case, the wrong prediction of the \texttt{restaurant-pricerange} slot at turn 0 will accumulate to the last turn. However, in this case, another incorrect prediction at turn 3 will cause error accumulation in this dialogue.

The second dialogue presented in Table \ref{table:dialog_2}, reports the incorrect prediction according to the interpretation of annotations at turn 4. In other words, because the dialogue about the \texttt{hotel-internet} slot appears over turns 4 and 5, it is solely an error depending on the prediction timing of the model. Because the correct belief state was predicted right from turn 5, it cannot be said to be an error accumulation phenomenon; however, the model did not predict the \texttt{hotel-pricerange} slot at turn 6, which is the last turn in this case.

In conclusion, it can be determined that the model does not seem to accumulate erroneous predictions because of an accidental situation or interpretation of annotations, but this does not negate the error accumulation phenomenon. Furthermore, the fact that the starting point of making the joint goal accuracy of subsequent turns to 0 mainly occurs at the beginning of the dialogue does not change.

\begin{table*}
    \setcounter{table}{0}
    \renewcommand{\thetable}{A\arabic{table}}
    \centering
    \begin{tabularx}{\textwidth}{c>{\raggedright}X}
        \toprule
        \textbf{Turn} & \textbf{Dialogue History} \tabularnewline
        \midrule
        \multirow{2}{*}{0} & \textit{System}: `` '' \\
            \textit{User}: ``can you help me find a nice restaurant ?'' \tabularnewline
        \midrule
        \multirow{2}{*}{1} & \textit{System}: ``sure ! what kind of food do you like ?'' \\
            \textit{User}: ``i was thinking some \textbf{indian} food would be great .'' \tabularnewline
        \midrule
        \multirow{2}{*}{2} & \textit{System}: ``i have 22 indian restaurant -s do you have a preference for area of town ?'' \\
            \textit{User}: ``no , i \textbf{do not care} where it is . i am thinking i would like an \textbf{expensive} restaurant .'' \tabularnewline
        \midrule
        \multirow{3}{*}{3} & \textit{System}: ``would you like to try \textbf{curry garden} ?'' \\
            \textit{User}: ``that is fine book me a table for \textbf{6} on \textbf{sat} at \textbf{17:30} . i also need a \textbf{train for the same} and should \textbf{leave leicester} for \textbf{cambridge}'' \tabularnewline
        \midrule
        \multirow{4}{*}{4} & \textit{System}: ``alright , i have made your requested booking at curry garden , and the reference number is hk9ycl6z . as for the train , what time would you like to depart ?'' \\
            \textit{User}: ``it does not matter as long as i am there by \textbf{13:45} leaving leicester going to cambridge , i'll need the reference number too please'' \tabularnewline
        \midrule
        \multirow{3}{*}{5} & \textit{System}: ``i have found tr6210 leaving leicester at 11:09 on saturday and arriving in cambridge at 12:54 . i have booked 6 seats . reference number is lr5i1rzv . anything else i can do ?'' \\
            \textit{User}: ``that will be all for today . thanks so much for all your help . goodbye'' \tabularnewline
        \bottomrule
    \end{tabularx}
    \caption{Sample dialogue of MultiWOZ 2.1 test set (PMUL4234.json).}
    \label{table:dialog_1}
\end{table*}

\begin{table*}
    \renewcommand{\thetable}{A\arabic{table}}
    \centering
    \begin{tabularx}{\textwidth}{c>{\raggedright}X}
        \toprule
        \textbf{Turn} & \textbf{Dialogue History} \tabularnewline
        \midrule
        \multirow{2}{*}{0} & \textit{System}: `` '' \\
            \textit{User}: ``i would like help finding a train headed to \textbf{cambridge} .'' \tabularnewline
        \midrule
        \multirow{3}{*}{1} & \textit{System}: ``i will be happy to help you find a train . can you tell me where you will be departing from ?'' \\
            \textit{User}: ``departing from \textbf{london kings cross} on \textbf{tuesday} .'' \tabularnewline
        \midrule
        \multirow{2}{*}{2} & \textit{System}: ``when would you like to leave or arrive by ?'' \\
            \textit{User}: ``i need to \textbf{arrive by 18,30} .'' \tabularnewline
        \midrule
        \multirow{2}{*}{3} & \textit{System}: ``take train tr1434 , which will arrive at 18:08 . shall i book you for that train ?'' \\
            \textit{User}: ``can i get the price for a ticket , first ?'' \tabularnewline
        \midrule
        \multirow{3}{*}{4} & \textit{System}: ``sure ! the ticket is 23.60 pounds .'' \\
            \textit{User}: ``thanks ! i am also looking for a hotel called \textbf{archway house} . can you tell me if they have \textbf{free wifi} ?'' \tabularnewline
        \midrule
        \multirow{2}{*}{5} & \textit{System}: ``they do . would you like to book a room ?'' \\
            \textit{User}: ``i would first like to know what their price range and hotel type are , thank you .'' \tabularnewline
        \midrule
        \multirow{3}{*}{6} & \textit{System}: ``archway house is a \textbf{moderate} -ly priced guesthouse . would you like their address or perhaps to book a room there ?'' \\
            \textit{User}: ``thank you , but no . you've already helped me with everything i needed today .'' \tabularnewline
        \bottomrule
    \end{tabularx}
    \caption{Sample dialogue of MultiWOZ 2.1 test set (MUL2270.json).}
    \label{table:dialog_2}
\end{table*}

\begin{table*}
\centering
\renewcommand{\thetable}{A\arabic{table}}
\begin{tabular}{cllc}
\toprule
\textbf{Turn} & \textbf{Predicted State} & \textbf{Gold State} & \textbf{Joint Goal Acc.}\\
\midrule
0 & restaurant-pricerange-\textcolor{red}{expensive} & -  & 0 \\ \hline
\multirow{2}{*}{1} & restaurant-pricerange-\textcolor{red}{expensive} & restaurant-food-indian & \multirow{2}{*}{0} \\ 
                   & restaurant-food-indian &  \\ \hline
\multirow{3}{*}{2} & restaurant-pricerange-expensive & restaurant-pricerange-expensive & \multirow{3}{*}{1} \\ 
                   & restaurant-food-indian & restaurant-food-indian \\ 
                   & restaurant-area-dontcare & restaurant-area-dontcare \\ \hline
\multirow{11}{*}{3} & restaurant-pricerange-expensive & restaurant-pricerange-expensive & \multirow{11}{*}{0} \\ 
                   & restaurant-food-indian & restaurant-food-indian \\ 
                   & restaurant-area-dontcare & restaurant-area-dontcare \\
                   & restaurant-book day-\textcolor{red}{sunday} & restaurant-book day-\textcolor{red}{saturday} \\
                   & restaurant-book people-6 & restaurant-book people-6 \\
                   & restaurant-book time-17:30 & restaurant-book time-17:30 \\
                   & restaurant-name-curry garden & restaurant-name-curry garden \\
                   & train-destination-cambridge & train-destination-cambridge \\
                   & train-day-\textcolor{red}{tuesday} & train-day-\textcolor{red}{saturday} \\ 
                   & train-departure-leicester & train-departure-leicester \\
                   &  & train-book people-\textcolor{red}{6} \\ \hline
\multirow{12}{*}{4} & restaurant-pricerange-expensive & restaurant-pricerange-expensive & \multirow{12}{*}{0} \\ 
                   & restaurant-food-indian & restaurant-food-indian \\ 
                   & restaurant-area-dontcare & restaurant-area-dontcare \\
                   & restaurant-book day-\textcolor{red}{sunday} & restaurant-book day-\textcolor{red}{saturday} \\
                   & restaurant-book people-6 & restaurant-book people-6 \\
                   & restaurant-book time-17:30 & restaurant-book time-17:30 \\
                   & restaurant-name-curry garden & restaurant-name-curry garden \\
                   & train-destination-cambridge & train-destination-cambridge \\
                   & train-day-\textcolor{red}{tuesday} & train-day-\textcolor{red}{saturday} \\ 
                   & train-departure-leicester & train-departure-leicester \\
                   & train-arriveby-13:45 & train-arriveby-13:45 \\
                   & train-leaveat-\textcolor{red}{dontcare} & train-book people-\textcolor{red}{6} \\ \hline
\multirow{12}{*}{5} & restaurant-pricerange-expensive & restaurant-pricerange-expensive & \multirow{12}{*}{0}\\ 
                   & restaurant-food-indian & restaurant-food-indian \\ 
                   & restaurant-area-dontcare & restaurant-area-dontcare \\
                   & restaurant-book day-\textcolor{red}{sunday} & restaurant-book day-\textcolor{red}{saturday} \\
                   & restaurant-book people-6 & restaurant-book people-6 \\
                   & restaurant-book time-17:30 & restaurant-book time-17:30 \\
                   & restaurant-name-curry garden & restaurant-name-curry garden \\
                   & train-destination-cambridge & train-destination-cambridge \\
                   & train-day-\textcolor{red}{tuesday} & train-day-\textcolor{red}{saturday} \\ 
                   & train-departure-leicester & train-departure-leicester \\
                   & train-arriveby-13:45 & train-arriveby-13:45 \\
                   & train-leaveat-\textcolor{red}{dontcare}  & train-book people-\textcolor{red}{6} \\
\bottomrule
\end{tabular}
\caption{\label{table:dialog_state_1}
SOM-DST prediction of MultiWOZ 2.1 test sample (PMUL4234.json).
}
\end{table*}

\begin{table*}
\centering
\renewcommand{\thetable}{A\arabic{table}}
\begin{tabular}{cllc}
\toprule
\textbf{Turn} & \textbf{Predicted State} & \textbf{Gold State} & \textbf{Joint Goal Acc.}\\
\midrule
0 & train-destination-cambridge & train-destination-cambridge & 1 \\ \hline
\multirow{3}{*}{1} & train-destination-cambridge & train-destination-cambridge & \multirow{3}{*}{1} \\ 
                   & train-day-tuesday & train-day-tuesday  \\ 
                   & train-departure-london kings cross & train-departure-london kings cross \\ \hline
\multirow{4}{*}{2} & train-destination-cambridge & train-destination-cambridge & \multirow{4}{*}{1}\\ 
                   & train-day-tuesday & train-day-tuesday  \\ 
                   & train-departure-london kings cross & train-departure-london kings cross \\ 
                   & train-arriveby-18:30 & train-arriveby-18:30 \\ \hline
\multirow{4}{*}{3} & train-destination-cambridge & train-destination-cambridge & \multirow{4}{*}{1} \\ 
                   & train-day-tuesday & train-day-tuesday  \\ 
                   & train-departure-london kings cross & train-departure-london kings cross \\ 
                   & train-arriveby-18:30 & train-arriveby-18:30 \\ \hline
\multirow{6}{*}{4} & train-destination-cambridge & train-destination-cambridge & \multirow{6}{*}{0} \\ 
                   & train-day-tuesday & train-day-tuesday  \\ 
                   & train-departure-london kings cross & train-departure-london kings cross \\ 
                   & train-arriveby-18:30 & train-arriveby-18:30 \\ 
                   & hotel-name-archway house & hotel-name-archway house \\ 
                   & & hotel-internet-\textcolor{red}{yes} \\ \hline
\multirow{6}{*}{5} & train-destination-cambridge & train-destination-cambridge & \multirow{6}{*}{1} \\ 
                   & train-day-tuesday & train-day-tuesday  \\ 
                   & train-departure-london kings cross & train-departure-london kings cross \\ 
                   & train-arriveby-18:30 & train-arriveby-18:30 \\ 
                   & hotel-name-archway house & hotel-name-archway house \\ 
                   & hotel-internet-yes & hotel-internet-yes \\ \hline
\multirow{7}{*}{6} & train-destination-cambridge & train-destination-cambridge & \multirow{7}{*}{0} \\ 
                   & train-day-tuesday & train-day-tuesday  \\ 
                   & train-departure-london kings cross & train-departure-london kings cross \\ 
                   & train-arriveby-18:30 & train-arriveby-18:30 \\ 
                   & hotel-name-archway house & hotel-name-archway house \\ 
                   & hotel-internet-yes & hotel-internet-yes \\ 
                   & & hotel-pricerange-\textcolor{red}{moderate} \\
\bottomrule
\end{tabular}
\caption{\label{table:dialog_state_2}
SOM-DST prediction of MultiWOZ 2.1 test sample (MUL2270.json).
}
\end{table*}

\begin{table*}
\centering
\renewcommand{\thetable}{A\arabic{table}}
\begin{tabular}{lll}
\toprule
\textbf{Method} & \textbf{Metric} & \textbf{Dataset}\\
\midrule
DST-STAR \citep{10.1145/3442381.3449939} & JGA & MultiWOZ 2.0 \citep{budzianowski-etal-2018-multiwoz}, \\
&& MultiWOZ 2.1 \citep{DBLP:journals/corr/abs-1907-01669} \\ 
Seq2Seq-DU \citep{feng-etal-2021-sequence} & JGA & SGD \citep{rastogi2020towards}, MultiWOZ 2.1,\\
&& MultiWOZ 2.2 \citep{zang-etal-2020-multiwoz}\\
L4P4K2-DSGraph \citep{lin2021knowledgeaware} & JGA, SA & MultiWOZ 2.0 \\
Transformer-DST \citep{zeng2021jointly} & JGA & MultiWOZ 2.0, MultiWOZ 2.1 \\
NA-DST \citep{Le2020Non-Autoregressive} & JGA, SA & MultiWOZ 2.0, MultiWOZ 2.1 \\
TripPy \citep{heck-etal-2020-trippy} & JGA & WOZ 2.0 \citep{wen-etal-2017-network}, MultiWOZ 2.1, \\
&& Sim-M, Sim-R \citep{shah2018building}\\
SOM-DST \citep{kim-etal-2020-efficient} & JGA, SA & MultiWOZ 2.0, MultiWOZ 2.1 \\
Simple-TOD \citep{NEURIPS2020_e9462095} & JGA & MultiWOZ 2.0, MultiWOZ 2.1 \\
GCDST \citep{wu-etal-2020-gcdst} & JGA & MultiWOZ 2.0, MultiWOZ 2.1 \\
CSFN-DST \citep{zhu-etal-2020-efficient} & JGA & MultiWOZ 2.0, MultiWOZ 2.1 \\
SAVN \citep{wang-etal-2020-slot} & JGA, SA & MultiWOZ 2.0, MultiWOZ 2.1 \\
SST \citep{Chen2020SchemaGuidedMD} & JGA, SA & MultiWOZ 2.0, MultiWOZ 2.1 \\
DS-DST \citep{zhang-etal-2020-find} & JGA & MultiWOZ 2.0, MultiWOZ 2.1 \\
DSTQA \citep{DBLP:journals/corr/abs-1911-06192} & JGA, SA & WOZ 2.0, MultiWOZ 2.0, MultiWOZ 2.1 \\
SUMBT \citep{lee-etal-2019-sumbt} & JGA & WOZ 2.0, MultiWOZ 2.0 \\
DST-Reader \citep{gao-etal-2019-dialog} & JGA & MultiWOZ 2.0 \\
BERT-DST \citep{DBLP:conf/interspeech/ChaoL19} & JGA & WOZ 2.0, Sim-M, Sim-R \\
&& DSTC2 \citep{henderson2014second}\\
TRADE \citep{wu-etal-2019-transferable} & JGA, SA & MultiWOZ 2.0 \\
HyST \citep{DBLP:conf/interspeech/GoelPH19} & JGA & MultiWOZ 2.0 \\
COMER \citep{ren-etal-2019-scalable} & JGA & WOZ 2.0, MultiWOZ 2.0 \\
\bottomrule
\end{tabular}
\caption{\label{table:survey}
Evaluation metrics used for performance comparison among the methodologies. We focused on metrics evaluating the belief state of each turn. For convenience, the name of each metric is abbreviated. JGA: Joint Goal Accuracy, SA: Slot Accuracy.
}
\end{table*}

\begin{table*}
\centering
\renewcommand{\thetable}{A\arabic{table}}
\setlength{\tabcolsep}{6.5pt}
\begin{tabular}{cp{5.1cm}p{5.1cm}lc}
\toprule
\multirow{2}{*}{\textbf{Turn}} & \multirow{2}{*}{\textbf{Predicted State}} & \multirow{2}{*}{\textbf{Gold State}} & \textbf{Slot} & \textbf{Relative} \\
&  &  &  \textbf{Acc.} & \textbf{Slot Acc.}\\
\midrule
0 &  \textcolor{red}{restaurant-name-nusha} & - & 0.9667 & 0 \\ \hline
1 &  \textcolor{red}{restaurant-name-nusha} & - & 0.9667 & 0  \\ \hline
2 &  \textcolor{red}{restaurant-name-nusha} &  \textcolor{red}{attraction-name-nusha} & 0.9333 & 0   \\ \hline
3 &  \textcolor{red}{restaurant-name-nusha} &  \textcolor{red}{attraction-name-nusha} & 0.9333 & 0  \\ \hline
\multirow{3}{*}{4} &  \textcolor{blue}{restaurant-area-centre} &  \textcolor{red}{attraction-name-nusha} & \multirow{3}{*}{0.9667} & \multirow{3}{*}{0.6667}  \\
  &  \textcolor{blue}{restaurant-food-indian} &  \textcolor{blue}{restaurant-area-centre} & &\\
  &  &  \textcolor{blue}{restaurant-food-indian} & & \\ \hline
\multirow{4}{*}{5} &  \textcolor{blue}{restaurant-area-centre} &  \textcolor{red}{attraction-name-nusha} & \multirow{4}{*}{0.9667} & \multirow{4}{*}{0.7500}  \\
  &  \textcolor{blue}{restaurant-food-indian} &  \textcolor{blue}{restaurant-area-centre} & & \\
  &  \textcolor{blue}{restaurant-pricerange-expensive} &  \textcolor{blue}{restaurant-food-indian} & & \\
  &  &  \textcolor{blue}{restaurant-pricerange-expensive} & & \\ \hline
\multirow{5}{*}{6} &  \textcolor{blue}{restaurant-name-saffron brasserie} &  \textcolor{red}{attraction-name-nusha} & \multirow{5}{*}{0.9667} & \multirow{5}{*}{0.8000}  \\
  &  \textcolor{blue}{restaurant-area-centre} &  \textcolor{blue}{restaurant-name-saffron brasserie} & & \\
  &  \textcolor{blue}{restaurant-food-indian} &  \textcolor{blue}{restaurant-area-centre} & & \\
  &  \textcolor{blue}{restaurant-pricerange-expensive} &  \textcolor{blue}{restaurant-food-indian} & & \\
  &  &  \textcolor{blue}{restaurant-pricerange-expensive} & & \\ \hline
\multirow{5}{*}{7} &  \textcolor{blue}{restaurant-name-saffron brasserie} &  \textcolor{red}{attraction-name-nusha} & \multirow{5}{*}{0.9667} & \multirow{5}{*}{0.8000}  \\
  &  \textcolor{blue}{restaurant-area-centre} &   \textcolor{blue}{restaurant-name-saffron brasserie} & & \\
  &  \textcolor{blue}{restaurant-food-indian} &  \textcolor{blue}{restaurant-area-centre} & & \\
  &  \textcolor{blue}{restaurant-pricerange-expensive} &  \textcolor{blue}{restaurant-food-indian} & &  \\
  &  &  \textcolor{blue}{restaurant-pricerange-expensive} & & \\ \hline
\multirow{5}{*}{8} &  \textcolor{blue}{restaurant-name-saffron brasserie} &  \textcolor{red}{attraction-name-nusha} & \multirow{5}{*}{0.9667} & \multirow{5}{*}{0.8000}  \\
  &  \textcolor{blue}{restaurant-area-centre} &  \textcolor{blue}{restaurant-name-saffron brasserie} & & \\
  &  \textcolor{blue}{restaurant-food-indian} &  \textcolor{blue}{restaurant-area-centre} & & \\
  &   \textcolor{blue}{restaurant-pricerange-expensive} &  \textcolor{blue}{restaurant-food-indian} & & \\
  &  &  \textcolor{blue}{restaurant-pricerange-expensive} & & \\ \hline
\multirow{5}{*}{9} &  \textcolor{blue}{restaurant-name-saffron brasserie} &  \textcolor{red}{attraction-name-nusha} & \multirow{5}{*}{0.9667} & \multirow{5}{*}{0.8000}  \\
  &  \textcolor{blue}{restaurant-area-centre} &  \textcolor{blue}{restaurant-name-saffron brasserie} & & \\
  &  \textcolor{blue}{restaurant-food-indian} &  \textcolor{blue}{restaurant-area-centre} & & \\
  &  \textcolor{blue}{restaurant-pricerange-expensive} &  \textcolor{blue}{restaurant-food-indian} & &  \\
  &  &  \textcolor{blue}{restaurant-pricerange-expensive} & & \\ 
\bottomrule
\end{tabular}
\caption{\label{table:extreme}
SOM-DST prediction of MultiWOZ 2.1 test sample (PMUL4648.json). The joint goal accuracy of every turn is 0 because of belief states with red color. When calculating score, the number of total slots is set to 30, which is of \textit{hotel, train, restaurant, attraction,} and \textit{taxi} domains in MultiWOZ 2.1. Relative slot accuracy can be calculated just using slot-values appearing in the dialogue, not being affected by unused information.
}
\end{table*}

\begin{figure*}[t!]
    \setcounter{figure}{0}
    \renewcommand{\thefigure}{A\arabic{figure}}
    \centering
    \renewcommand{\thetable}{A\arabic{table}}
    \begin{subfigure}[b]{0.32\textwidth}  
        \centering 
        \includegraphics[width=\textwidth]{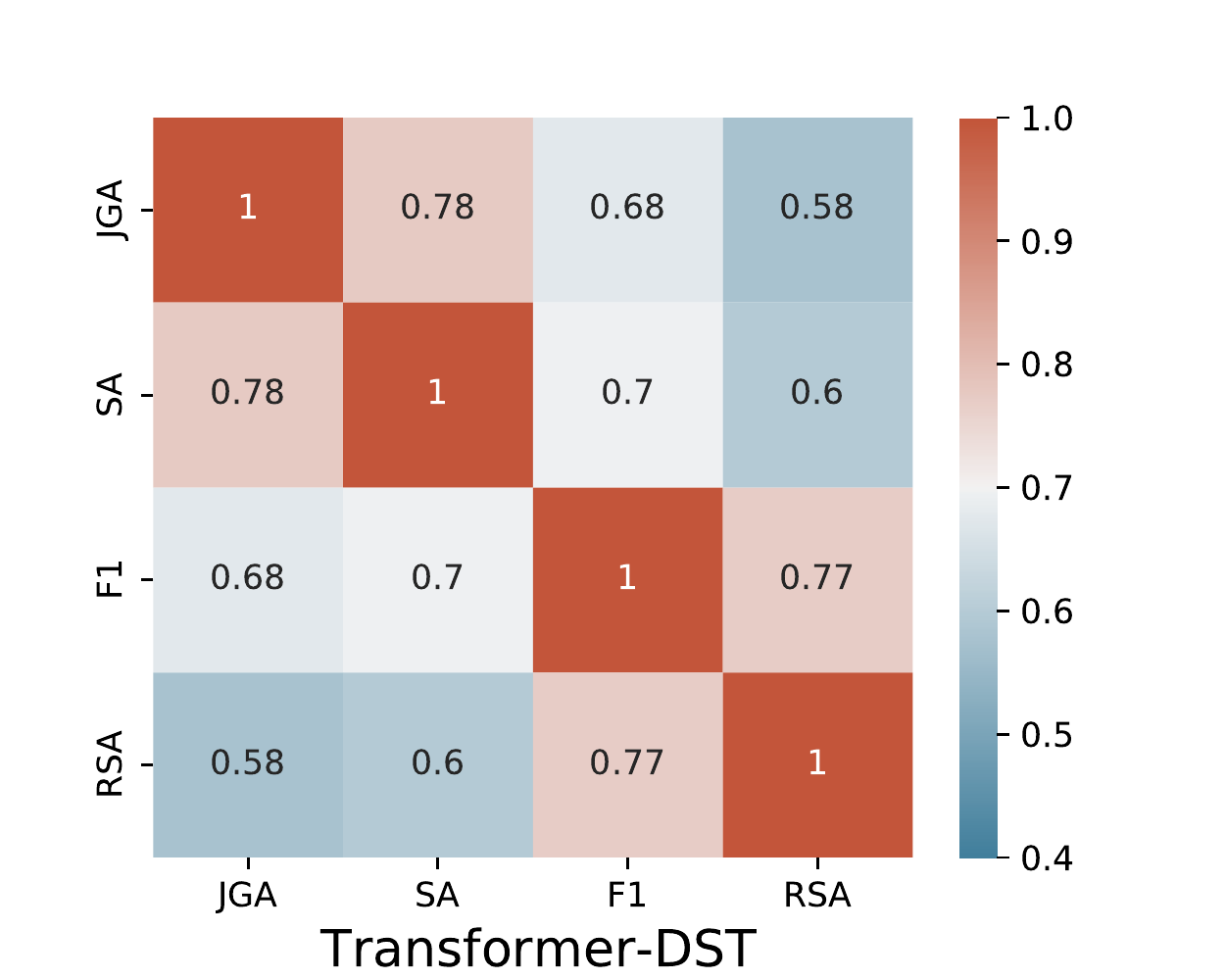}
    \end{subfigure}
    \begin{subfigure}[b]{0.32\textwidth}  
        \centering 
        \includegraphics[width=\textwidth]{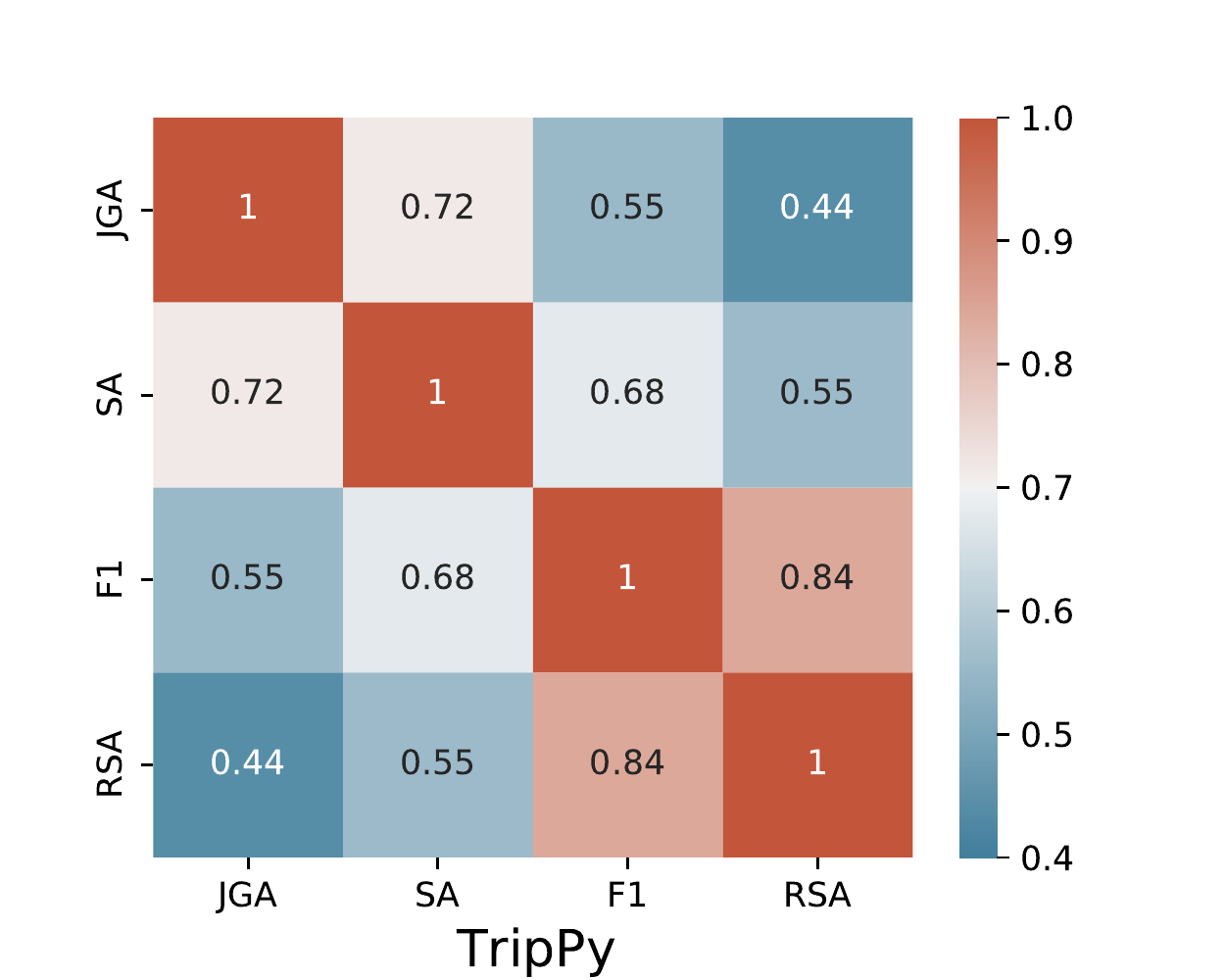}
    \end{subfigure}
    \begin{subfigure}[b]{0.32\textwidth}  
        \centering 
        \includegraphics[width=\textwidth]{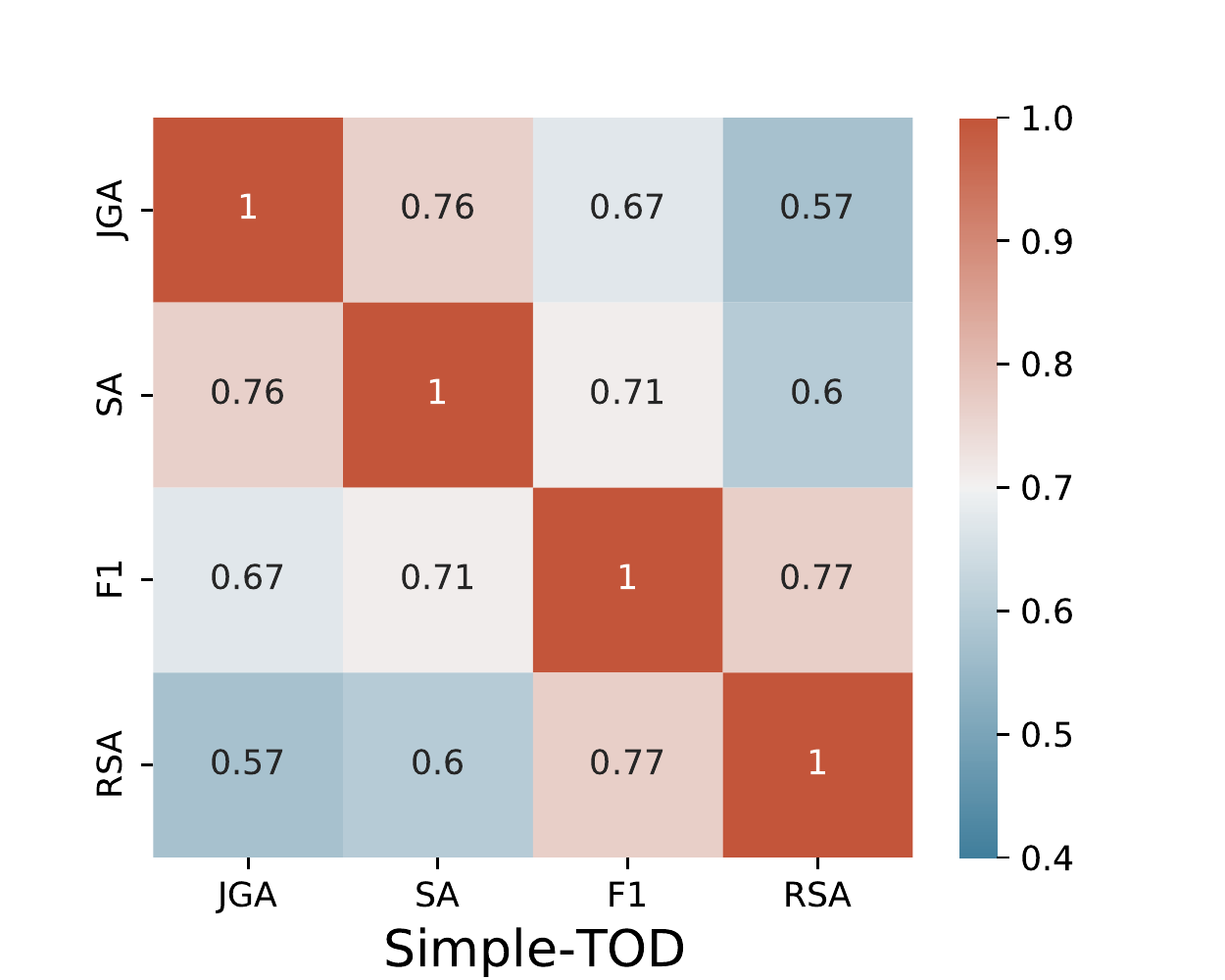}
    \end{subfigure}
    \begin{subfigure}[b]{0.32\textwidth}  
        \centering 
        \includegraphics[width=\textwidth]{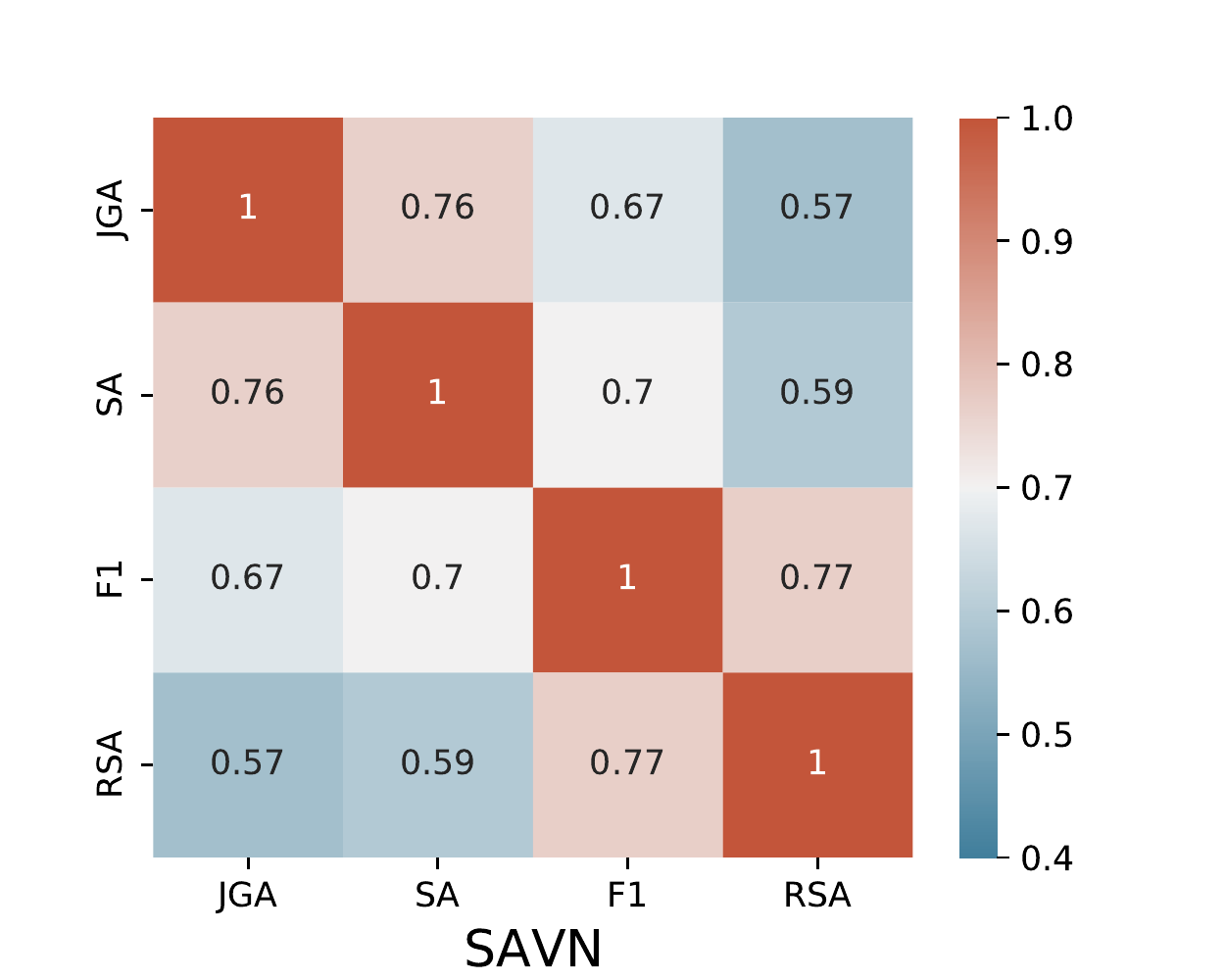}
    \end{subfigure}
    \begin{subfigure}[b]{0.32\textwidth}
        \centering
        \includegraphics[width=\textwidth]{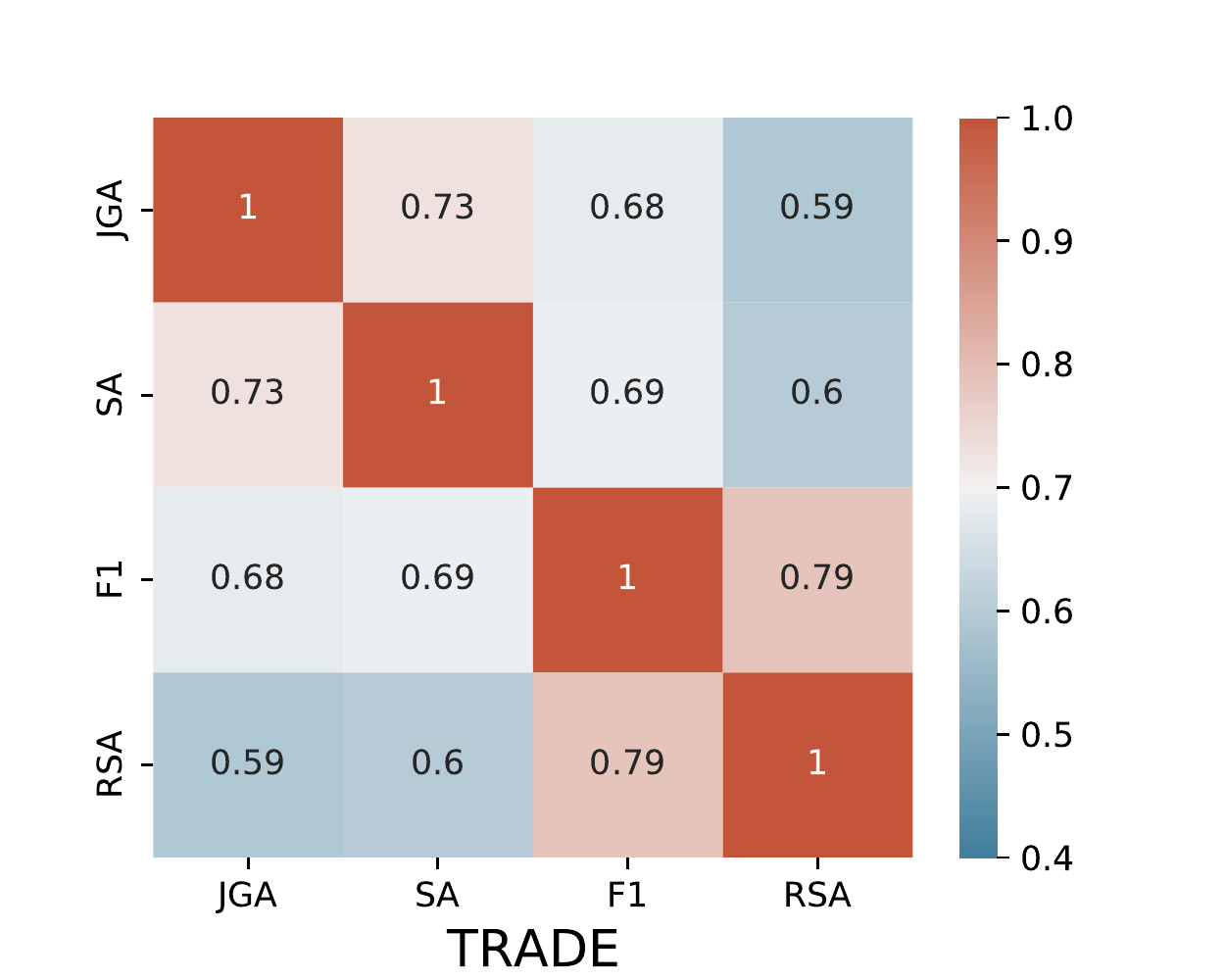}
    \end{subfigure}
    \begin{subfigure}[b]{0.32\textwidth}  
        \centering 
        \includegraphics[width=\textwidth]{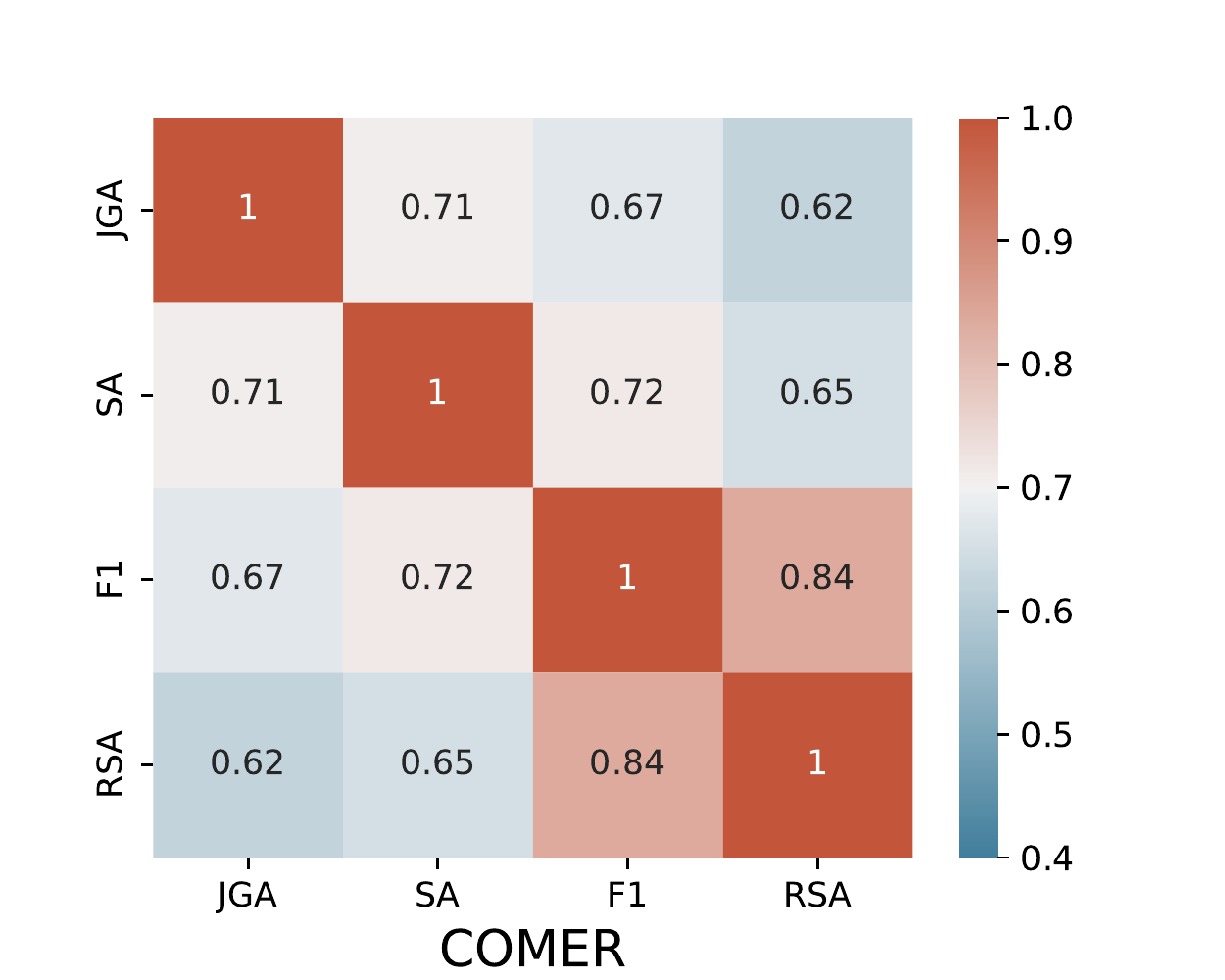}
    \end{subfigure}
    \begin{subfigure}[b]{0.32\textwidth}  
        \centering 
        \includegraphics[width=\textwidth]{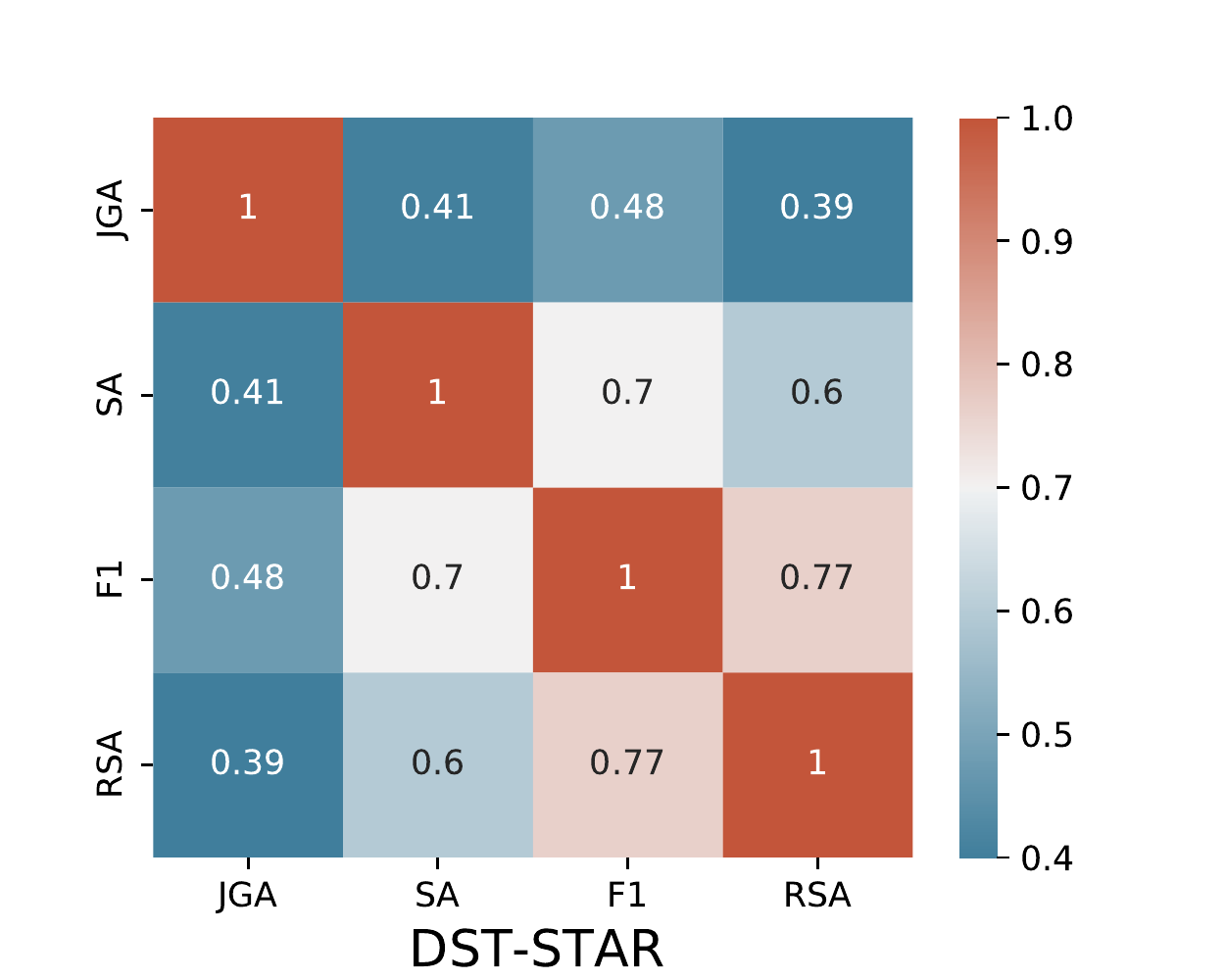}
    \end{subfigure}
    \begin{subfigure}[b]{0.32\textwidth}  
        \centering 
        \includegraphics[width=\textwidth]{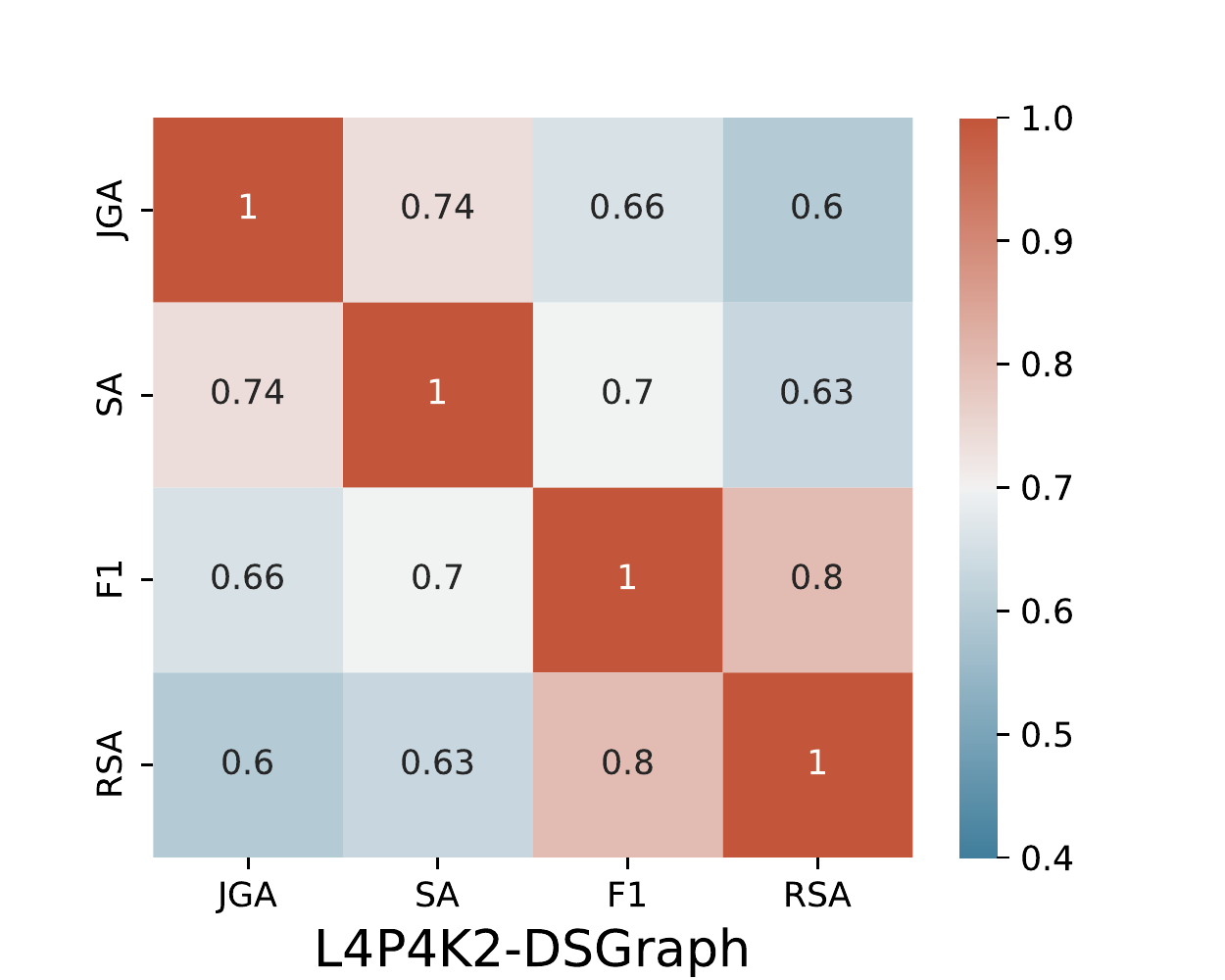}
    \end{subfigure}
    \begin{subfigure}[b]{0.32\textwidth}  
        \centering 
        \includegraphics[width=\textwidth]{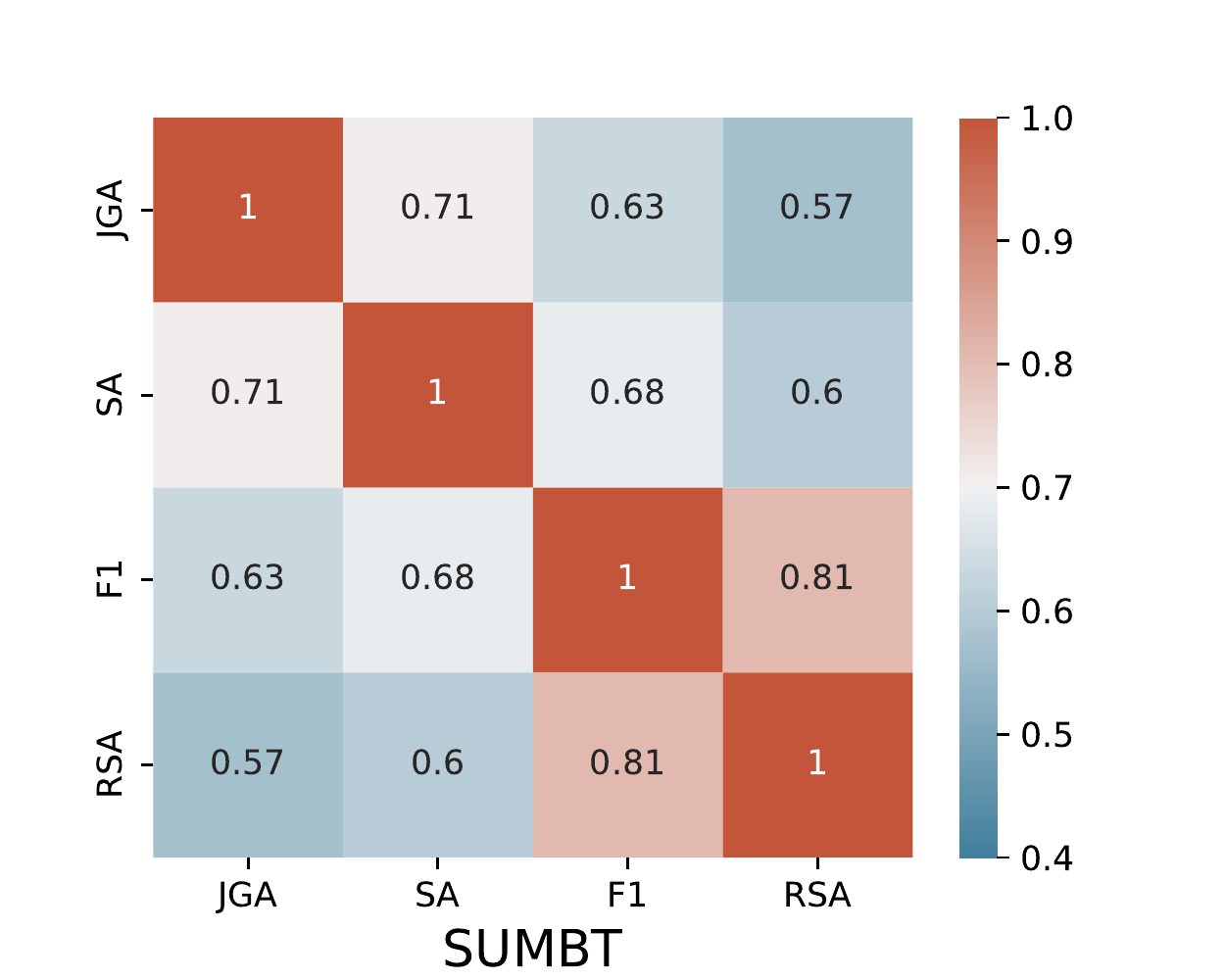}
    \end{subfigure}
    \caption
    {Correlation matrices of evaluation performance using various DST models.} 
    \label{fig:corrs}
\end{figure*}

\end{document}